\pdfoutput=1

\documentclass[11pt]{article}
\PassOptionsToPackage{table,xcdraw}{xcolor}
\usepackage{graphicx}
\usepackage[review]{ACL2023}

\usepackage{times}
\usepackage{latexsym}

\usepackage{arydshln}
\usepackage{amsmath}
\usepackage{relsize}
\usepackage{enumitem}
\usepackage[T1]{fontenc}

\usepackage[utf8]{inputenc}

\usepackage{microtype}

\usepackage{inconsolata}

\usepackage{booktabs}
\usepackage{multirow}

\usepackage{setspace}

\title{CodeTool: Enhancing Programmatic Tool Invocation of LLMs \\via Process Supervision}

\author{%
  Yifei Lu$^{1,2}$$^{\ast}$, Fanghua Ye$^{2}$$^{\ast}$, Jian Li$^{2}$$^{\dagger}$, Qiang Gao$^{2}$, Cheng Liu$^{2}$,\\
  \textbf{Haibo Luo$^{1}$, Nan Du$^{2}$, Xiaolong Li$^{2}$, Feiliang Ren$^{1}$$^{\dagger}$}\\
  $^1$Northeastern University, Shenyang, China \\
  $^2$Hunyuan AI Digital Human, Tencent, Shenzhen, China \\
  \texttt{lyfei1126@gmail.com}, \texttt{\{fanghuaye, jackjianli\}@tencent.com} \\
  \texttt{renfeiliang@cse.neu.edu.cn}
}

\date{}
\begin{document}
\maketitle
\footnotetext[1]{Work done during an internship at Tencent Hunyuan.}
\begingroup
\renewcommand\thefootnote{}\footnotetext{$^{\ast}$ Equal Contribution.}
\renewcommand\thefootnote{}\footnotetext{$^{\dagger}$ Corresponding Author.}
\endgroup

\begin{abstract}

Tool invocation significantly enhances the capabilities of Large Language Models (LLMs), yet challenges persist, particularly in complex task scenarios. Current methods, such as instruction-enhanced reasoning and supervised fine-tuning, often result in unnecessarily long reasoning depths and face difficulties in verifying the correctness of intermediate steps. In this paper, we propose \textbf{CodeTool}, a novel framework for stepwise code generation that improves LLM tool invocation by leveraging the concise and easily verifiable nature of code. CodeTool incorporates two distinct process rewards: the \textbf{On-the-spot Reward}, which provides immediate feedback on the accuracy of each tool invocation, and the \textbf{Latent Reward}, which assesses the contribution of each step toward overall task completion. By maximizing the cumulative reward of the On-the-spot and Latend Rewards at each step, LLMs are guided to follow efficient and accurate reasoning paths. Extensive experiments on StableToolBench and RestBench-TMDB demonstrate the superiority of CodeTool over existing approaches. Our implementation is available at \href{https://github.com/LimOkii/CodeTool}{LimOkii/CodeTool}.



\end{abstract}

\begin{figure*}[t]
    \centering
    \includegraphics[width=1\linewidth]{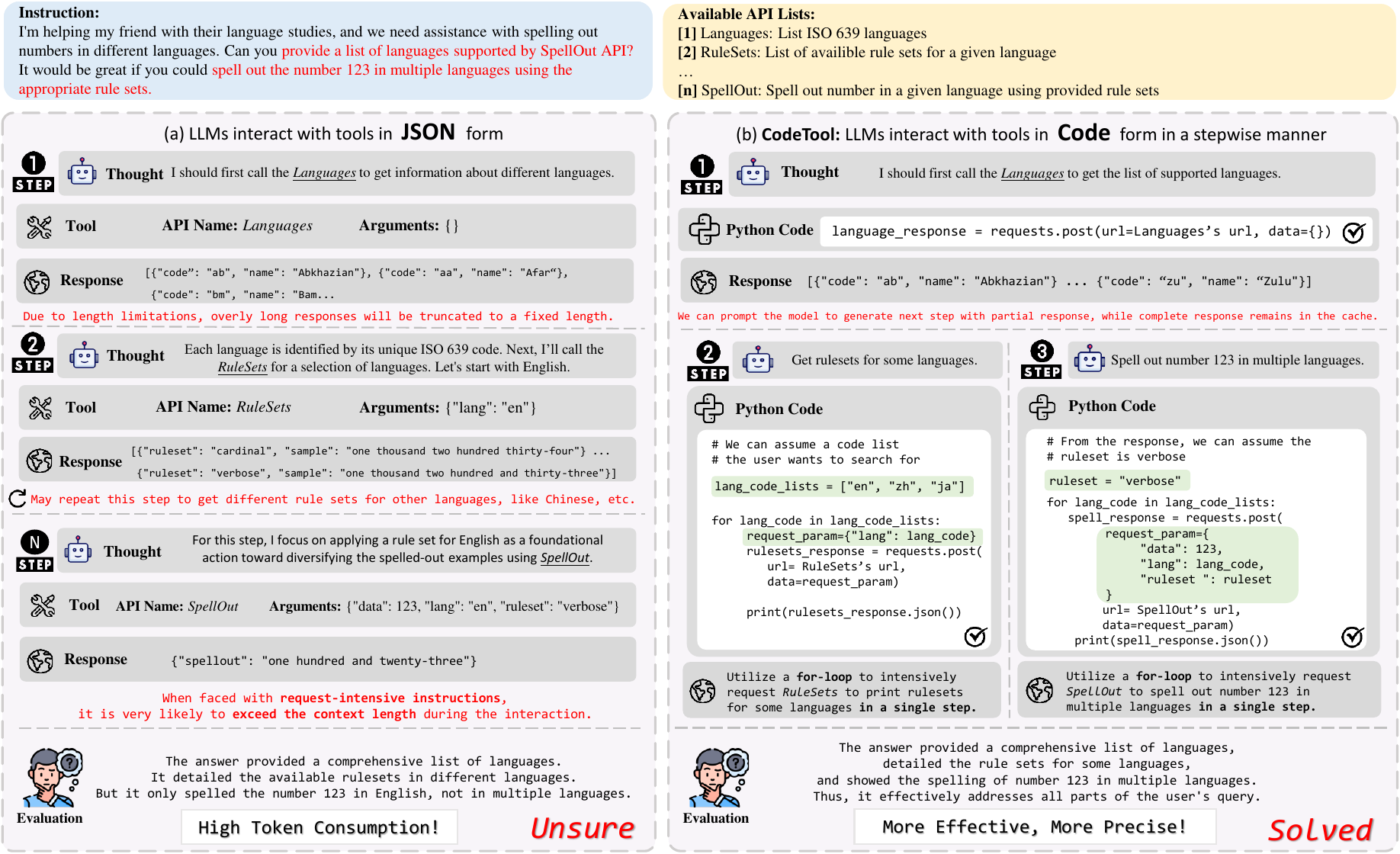}
    \caption{Comparison of tool invocation in JSON format and our proposed stepwise code generation framework: (a) JSON-based invocation is token-heavy when handling request-intensive tasks, and is prone to truncation, risking the loss of key information. (b) The stepwise code generation framework uses loops to handle request-intensive tasks efficiently, with stepwise supervision ensuring correctness of intermediate steps.}
    \label{fig:overview}
    \vspace{-0.1in}
\end{figure*}

\section{Introduction}

Tool invocation grants Large Language Models (LLMs) the ability to access external tools \citep{toolformer, hugginggpt, toollearningfoundationmodels}, thereby significantly expanding their range of capabilities. Despite the strong potential and ability demonstrated by LLMs in various tasks \citep{fewshot, gpt4_report}, they still encounter challenges when performing tool invocations in complex scenarios.

Early studies \citep{cot, react, restgpt} have assisted LLMs in better tool invocation by enabling them to think step by step or through instruction enhancement. While this approach is straightforward, it fails to fully leverage the potential of LLMs. More recent studies \citep{toolllama, toolalpaca, gorilla} have sought to enhance the tool invocation capabilities of LLMs via Supervised Fine-Tuning (SFT). However, training models on static trajectories of successful executions through text generation constrains their adaptability to novel tasks and environments. Besides, these existing studies primarily focus on tool invocation in Text or JSON format, which often leads to prolonged reasoning paths.

Programmatic tool invocation offers a more flexible and generalizable alternative to Text or JSON-based approaches \citep{describeexplain, codeact}. By leveraging programming constructs such as loops (\textit{for} or \textit{while}) and arrays, code can efficiently handle request-intensive instructions, thereby reducing the number of interactions required, as illustrated in Figure \ref{fig:overview}. 
However, existing code-based approaches still face two key challenges. \textit{First}, relying solely on the built-in functions of programming languages and a limited set of predefined libraries \citep{codeact} restricts the quantity and scope of tools that can be invoked. \textit{Second}, generating complete code in a single pass \citep{ACT}, albeit increasing the number of accessible tools, lacks supervision over intermediate steps, making it difficult to detect and correct errors in complex scenarios. 

Supervising the correctness of intermediate steps (i.e., process supervision) during an LLM's reasoning process has been shown to improve the final accuracy of challenging tasks \citep{solvingmathwordproblems, letsverifystepstep, mathshepherd}. This approach, however, typically requires large-scale annotations of process data. In addition, the supervision signal tends to direct the model toward plausible correct answers rather than ensuring absolute correctness \citep{prime}. 
A recent study by \citet{steptool} attempts to address these challenges by incorporating process rewards within a reinforcement learning framework to enhance tool invocation. While this method holds promise, its rewards are simply generated by strong LLMs, raising concerns about its objectivity and reliability.

In this work, we propose \textbf{CodeTool}, a novel stepwise code generation framework designed to enhance tool invocation of LLMs. CodeTool introduces two distinct process rewards during inference: the \textbf{On-the-spot Reward} and the \textbf{Latent Reward}. The On-the-spot Reward leverages the inherently verifiable nature of code to provide immediate feedback on the correctness of each tool invocation, ensuring precise execution at every step. In contrast, the Latent Reward, assigned by a trained Process Reward Model (PRM), evaluates the potential contribution of each step towards the overall task completion.   At each step, LLMs are guided to follow the reasoning direction that maximizes the cumulative reward of the On-the-spot and Latent Rewards, as illustrated in Figure \ref{fig:illustrate}.
This dual-reward mechanism overcomes key challenges in current programmatic approaches, particularly the lack of supervision over intermediate reasoning steps. Moreover, the On-the-spot Reward, which is grounded in the executability of the generated code, ensures objective and highly reliable feedback, as demonstrated by our experiments in Section \ref{sec:main_res_ans}.

The contributions of this work are as follows:
\begin{itemize}[leftmargin=0.4cm]
    \item We propose CodeTool, a stepwise code generation framework that leverages process supervision to enhance the capabilities of LLMs in tool invocation.
    \item We design two types of process rewards--\textit{On-the-spot Reward} and \textit{Latent Reward}--to provide high-quality process supervision, considering both immediate feedback and long-term potential.
    \item We conduct extensive experiments on StableToolBench \citep{stabletoolbench} and RestBench-TMDB \citep{restgpt}, confirming the superiority of CodeTool over existing methods.
\end{itemize}

\section{Related Work}
\subsection{Tool Invocation With LLMs}
Previous studies \citep{cot, react, restgpt} have investigated enabling LLMs to interact with various tools, such as search engines, calculators, translation software, and third-party API services, to facilitate tool utilization. Most of these approaches rely on prompt engineering to enhance the reasoning capabilities of LLMs during inference or to design prompts tailored to specific modules and tools. Subsequent research \citep{toolllama, toolalpaca} has focused on fine-tuning open-source LLMs to equip them with the ability to invoke tools. Recognizing that tool invocation in complex scenarios often requires multi-step reasoning, some studies \citep{tpllama, wang2024trialanderror} have shifted attention toward how LLMs can learn to use tools effectively from error-prone calls. Additionally, \citet{steptool} has explored the integration of reward mechanisms in the intermediate decision-making process of LLMs, employing reinforcement learning techniques to improve tool invocation efficiency and outcomes.

\subsection{Programmatic Tool Invocation}
LLMs typically generate action units in pre-defined formats (e.g., JSON or Text) to interact with external tools. In contrast, programmatic tool invocation offers an alternative mode of interaction. Recent studies have highlighted the potential of incorporating programming to enhance the planning and reasoning capabilities of LLMs, with the feasibility of code-based reasoning particularly demonstrated in complex numerical reasoning tasks \citep{programthoughts, programaidedllm}. Within the context of tool invocation, code blocks can be considered as action units for requesting or executing specific tools. For instance, \citet{codeact} and \citet{ACT} have investigated how LLMs can generate complete code to invoke Python's built-in functions or access third-party API services, thereby addressing intricate user instructions. However, these approaches often neglect the significant impact that the accuracy of intermediate steps can have on the final outcome. Notably, code data has been integrated into LLM pretraining \citep{codellama, wizardcoder, qwen_coder, deepseek_coder}, resulting in models that demonstrate advanced proficiency in structured programming, thus facilitating the cost-effective adoption of programmatic tool invocation.

\subsection{Process Supervision Methods}
While LLMs exhibit impressive capabilities across a wide range of tasks, they continue to encounter difficulties in reasoning through complex problems.  \citet{letsverifystepstep} has shown that supervising the correctness of intermediate steps in reasoning tasks can significantly improve the likelihood of LLMs producing accurate final answers. \citet{mathshepherd} and  \citet{improvemathematicalreasoninglanguage} have proposed automated approaches for constructing intermediate process data. However, their reward designs remain relatively simplistic, focusing solely on the potential of a given step to lead to a correct final answer, while neglecting the correctness of the step itself. \citet{steptool} has extended the reward framework to intermediate steps during tool invocation, but the acquisition of process rewards heavily relies on GPT-based annotations. In this work, we seek to fully automate the construction of a performant process reward system to improve programmatic tool invocation of LLMs.

\section{Methodology}
In this section, we first propose \textit{CodeTool}, a stepwise code generation framework to effectively address the challenges of tool invocation in complex scenarios. Subsequently, we design a process reward system to evaluate each decision-making step during tool invocation. Finally, we train a PRM on fully automated process data and rewards. The architecture of CodeTool is illustrated in Figure~\ref{fig:illustrate}.


\subsection{Preliminaries}
Addressing real-world user queries with the help of external tools can be conceptualized as a stepwise planning and reasoning process. Formally, let  \( \mathcal{M} \) represent an LLM with access to a set of real-world tools $\mathcal{T} = \{ t_1, t_2, \dots, t_{|\mathcal{T}|} \}$, where each tool \( t_i \) is associated with a logging protocol \( d_i \in \mathcal{D} = \{ d_1, d_2, \dots, d_{|\mathcal{D}|} \} \), which provides meta-information such as the tool’s description and the parameters required to make requests.


The goal of CodeTool is to iteratively write Python code  \( \mathcal{C}_{t} \) at each step to select the appropriate tool and issue requests with the correct parameters to obtain responses, ultimately deriving the final answer. Compared to previous work \cite{ACT} that solves user queries by directly generating complete code based on a well-structured dataset with detailed input-output schemas (which are hard to obtain in practice), the stepwise code generation method in CodeTool reduces reliance on such datasets. It allows the model to generate code to invoke a tool at each step, with the flexibility to utilize partial responses from previous invocations in subsequent steps. Such a sequential approach enables the model to comprehend the content and structure of intermediate outputs, thus aiding in parsing the responses and determining the next appropriate tool to invoke. Moreover, different from text-based responses, which are often truncated to a fixed length when they are overly long \cite{toolllama}, the results from code-based tool invocation are complete. The critical data is preserved in the cache, remaining intact and unaffected by truncation, thereby ensuring the integrity of the tool responses throughout the entire process.

\subsection{CodeTool: Stepwise Code Generation}
Given a user query \( q \), we first provide the LLM with the documented protocol \( d_i \in \mathcal{D} \) for each tool \( t_i \) in the candidate toolset \( \mathcal{T} \). Each protocol \( d_i \) contains meta-information, including a description of the purpose of the tool, the URL to be requested, and the argument requirements for invoking it. Then, we instruct the LLM \( \mathcal{M} \) to generate executable programs step by step to utilize multiple tools and ultimately solve the query \( q \).


Formally, this process can be formulated as:
\begin{equation}
    \mathcal{C}_t = \mathcal{M}(q, \mathcal{T}, \mathcal{D}, I_c, r_{t-1}), \quad r_{0}=\emptyset,
\end{equation}
where \( I_c \) indicates a concise instruction for stepwise code generation (refer to Appendix \ref{appendix:instruction} for details). The intermediate result \( r_t \) of the $t$-th step is obtained by executing the generated program $\mathcal{C}_t$ through a code interpreter, which is formulated as:
\begin{equation}
    r_t = \textsc{Execute}(\mathcal{C}_{t}).
\end{equation}
It is worth noting that there is initially no response from the tool, meaning that $ r_{0}=\emptyset. $  The subsequent program generation operations will parse specific field information based on the responses from the current tool and then invoke the next tool, continuing this process until the final answer is obtained.


\begin{figure}[!t]
    \centering
    \setlength{\abovecaptionskip}{2mm}
    \includegraphics[width=\columnwidth]{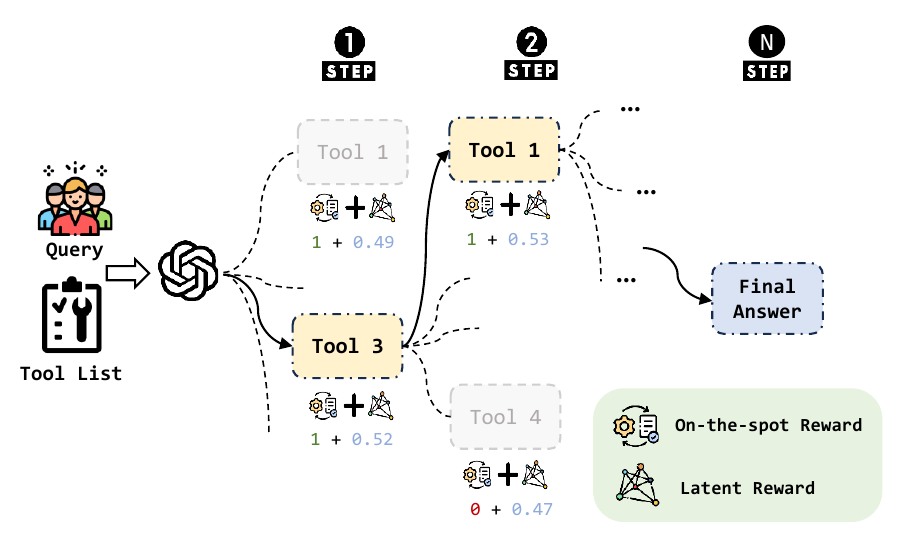}  
    \caption{The architecture of \textbf{CodeTool}, a stepwise code generation framework guided by two types of process rewards during inference. At each inference step, LLMs follow the reasoning path that maximizes the cumulative rewards of \textbf{\textit{On-the-spot Reward}} and \textbf{\textit{Latent Reward}}.}
    \label{fig:illustrate}
\end{figure}

\subsection{Process Reward Supervision}
\label{sec:reward}

In this part, we detail the integration of process supervision into the stepwise code generation framework to enhance its performance.
As illustrated in Figure \ref{fig:illustrate}, at each step, we first sample multiple candidate actions for the next step and then select the optimal action based on a calculated process reward to proceed to the next step.
This approach facilitates the exploration of a broader range of options, thereby increasing the likelihood of selecting the appropriate tool and generating accurate code for each step.
We consider two types of process rewards: \textit{On-the-spot Reward} and \textit{Latent Reward}.


\paragraph{On-the-spot Reward}
On-the-spot Reward evaluates whether the model has provided correct and executable code at the current step, including verifying whether a valid request body has been given within the candidate toolset or not. It can be obtained without any external supervision, as it only requires the automatic execution of the generated code \(\mathcal{C}_{t} \) using a Python interpreter \textsc{Execute}.

On-the-spot Reward is defined as:
\begin{equation}
R_{\text{spot}, t} = 
\begin{cases} 
1, & \text{if } \textsc{Execute}(\mathcal{C}_{t}) \text{ is successful}; \\
0, & \text{otherwise}.
\end{cases}
\end{equation}
This reward serves as a necessary condition to ensure that the model's reasoning moves along a potentially correct path. It provides immediate feedback on whether the generated code is executable and correct at each step. However, it only measures the correctness of the current step, ignoring whether the current step contributes to completing the user's query. Therefore, we also need to introduce another Latent Reward to assess the potential value of the current step in completing the query.

\paragraph{Latent Reward}
Latent Reward evaluates the potential value of the current step in helping the model successfully complete the task, considering factors such as whether redundant tools are invoked, leading to an unnecessarily long reasoning path, and whether the model selects an incorrect tool.


Drawing on previous approaches \citep{letsverifystepstep, mathshepherd, improvemathematicalreasoninglanguage}, we employ the Monte Carlo Tree Search (MCTS) algorithm to estimate the Latent Reward. Specifically, from each reasoning step, we expand the search tree based on a random sampling of the search space, resulting in multiple executed paths or rollouts. Then, the Latent Reward is defined as: 
\begin{equation}
\label{equa:raw_latent_reward}
    LR(q, s_{1:t}) = \frac{\Delta_{correct}}{\Delta_{total}},
\end{equation}
where $\Delta_{correct}$ and $\Delta_{total}$ denote the number of paths that are correctly executed with final answer labeled as \texttt{``Solved''} and the total number of executed paths, respectively, from the $t$-th step to its leaf nodes, while \(s_{1:t} \) denotes the reasoning sequence from the first step to step \(t\).


However, in complex scenarios, tool invocation may suffer from issues such as repetitive calls to a specific tool, particularly when the tool is no longer accessible, and long and redundant tool invocations. To mitigate unnecessary resource consumption and inefficiency, it is crucial to prioritize shorter, yet correct, tool invocation paths. Therefore, a penalty mechanism should be implemented to discourage such inefficiencies \cite{improvemathematicalreasoninglanguage}. In light of this, the final Latent Reward is defined as:
\begin{equation}
\label{equa:latent_reward}
    R_{\text{latent}, t}(q,s_{1:t}) =\alpha ^{1-LR\left( q, s_{1:t} \right)}\cdot \beta ^{\frac{\tau}{L}},
\end{equation}
where \(\alpha, \beta \in (0, 1]\) and \(L > 0\) are constant hyperparameters, while \(\tau\) denotes the average number of steps of executed paths starting from the $t$-th step. This formula comprehensively takes into account both the initial Latent Reward and the penalty for reasoning with overly long or redundant steps.


Based on the above two types of rewards, we obtain the cumulative reward of the $t$-th step:
\begin{equation}
    R_{\text{total}, t}(q,s_{1:t}) = R_{\text{spot}, t} + R_{\text{latent}, t}(q,s_{1:t}).
\end{equation}
Then, we select the candidate action with the highest cumulative reward to move to the next step.


\subsection{Process Latent Reward Model Training}
\label{sec:3.4}
During inference, while the On-the-spot Reward can be readily obtained through a code interpreter, estimating the Latent Reward using the MCTS algorithm presents significant challenges, including: (1) the time-intensive and costly nature of collecting multiple rollouts, and (2) the difficulty in evaluating whether a rollout successfully addresses the task in the absence of ground-truth data, a common scenario in practice. To address these challenges, we propose training a process Latent Reward model to estimate the Latent Reward during inference, significantly enhancing efficiency.

For model training, we select user queries from the  ToolBench \citep{toolllama} training set that remain solvable, meaning that the tools or APIs required to solve these queries are still callable, and use them to construct the intermediate process data.
To balance computational resource usage and inference time, we employ a depth-first search algorithm to construct an action search space resembling a binary tree for each user query. 
Specifically, during the code generation for each tool invocation step, we perform sampling twice.
Subsequently, at each hierarchical level of the action search space, we collect process data with varying Latent Reward values, which are regarded as a key indication for comparing the levels of potential. Ultimately, we leverage the data to train the PRM. The entire process is fully automated, ensuring efficiency and requiring no human intervention. Further details can be found in the Appendix \ref{appendix:details}

\begin{table}[!t]
    \centering
    \small
    \resizebox{\columnwidth}{!}{
    \begin{tabular}{lccccccc}
    \toprule
     & \textbf{I1-I}  & \textbf{I1-C} & \textbf{I1-T}& \textbf{I2-I} & \textbf{I2-C} & \textbf{I3-I} & \textbf{Total}\\
     \midrule
    \textbf{Full} & 200 & 200 & 200 & 200 & 200 & 100 & 1100 \\
    \textbf{Solvable} & 163 & 153 & 158 & 106 & 124 & 61 & 765 \\ 
    \textbf{Filtered} & 131 & 122 & 118 & 92 & 100 & 17 & 580 \\
    \bottomrule
    \end{tabular}}
    \caption{Statistics regarding the original full and solvable tasks provided by ToolBench and StableToolBench, along with the statistics on the data after our further screening. C, I, T stands for the `Category', `Instruction' and `Tool' subgroup of the test set, respectively.}
    \label{tab:data_stats}
\end{table}

\section{Experimental Setup}

\subsection{Datasets}

ToolBench \cite{toolllama} is a widely adopted benchmark in the field of tool invocation, designed to evaluate the ability of models to solve user instructions in complex real-world scenarios. As mentioned above, for the training of the process Latent Reward model, we select tools and APIs that are still accessible from the ToolBench training set and automatically construct intermediate process data for tool invocation. 
To evaluate the performance of CodeTool, we utilize the ToolBench test set. However, the original ToolBench test set presents challenges in reproducibility due to the inaccessibility of many tools and APIs.
To address this issue, we employ StableToolBench \citep{stabletoolbench}, a stable version of the Rapid-API access system derived from ToolBench using an API caching mechanism. StableToolBench comprises 765 solvable tasks distributed across six subsets, with each varying in tool category and instruction complexity, ranging from single-tool to multi-tool instructions. The detailed statistics are presented in Table \ref{tab:data_stats}.


\subsection{Adaptations to  StableToolBench}
\paragraph{API Response Handling}

In StableToolBench, API responses are provided in Text format, and excessively long responses are truncated to a fixed length. In order to facilitate the processing of these responses, we convert them to JSON format upon receipt, ensuring that the data can be effectively utilized by the code. For responses that are truncated, we employ GPT-4 (\texttt{GPT-4-Turbo-2024-04-09}) to reconstruct them into a complete JSON format. 

\paragraph{Test Set Filtering}
In our experiments, we find that even when the model provides the same API and request parameters as those used in the StableToolBench experiments, some requests fail due to the absence of cache hits. 
To ensure a fair comparison of experimental results, we filter out entries from the StableToolBench test set that are non-reproducible. The specific criteria for exclusion are detailed in Appendix \ref{appendix:rules}.


\subsection{Baselines and Evaluation Metric}
\label{sec:baseline_metric}
We conduct experiments using both open-source and closed-source LLMs.
For open-source LLMs, we primarily compare our method with the well-established baseline, \textbf{ToolLLaMA} \citep{toolllama}, which is fine-tuned from LLaMA on successful tool execution chains. Additionally, we include two reinforcement learning-based baselines, \textbf{TP-LLaMA} \citep{tpllama} and \textbf{StepTool} \citep{steptool}, which utilize direct preference optimization  \citep{dpo} and proximal policy optimization \citep{ppo}, respectively. 
To take full advantage of the code generation capability of LLMs and ensure a fair comparison at the same time, we employ \textbf{CodeLlama-7B}\footnote{\url{https://huggingface.co/codellama/CodeLlama-7b-hf}}, \textbf{Qwen2.5-7B-Instruct}\footnote{\url{https://huggingface.co/Qwen/Qwen2.5-7B-Instruct}}
and \textbf{Qwen2.5-Coder-7B-Instruct}\footnote{\url{https://huggingface.co/Qwen/Qwen2.5-Coder-7B-Instruct}} as the code generation models without any additional fine-tuning for CodeTool. All models share the same number of parameters (i.e., 7B) as the baselines.
For closed-source LLMs, we adopt \textbf{GPT-3.5-Turbo-0613} and \textbf{GPT-4-Turbo-Preview}, each representing different performance levels and capabilities within the GPT series.
Following \citet{toolllama, steptool}, we employ two inference strategies for baselines: 
Chain of Thought (\textbf{CoT}) \citep{cot} and Depth-First Search Decision Tree (\textbf{DFSDT}) \citep{toolllama}.

Following StableToolBench, we utilize the Solvable Pass Rate (\textbf{SoPR}) as the evaluation metric. Specifically, GPT-4 (\texttt{gpt-4-turbo-2024-04-09}) is leveraged as the evaluator to categorize the answers into \texttt{``Solved''}, \texttt{``Unsure''}, or \texttt{``Unsolved''}, with corresponding scores of 1, 0.5, and 0, respectively, contributing to the overall SoPR calculation. However, evaluation experiments on StableToolBench have shown that different models exhibit varying preferences regarding the degree to which the final answer resolves the query, leading to unstable evaluation results. To address this, we test the evaluation script provided by StableToolBench and introduce clearer criteria for assessing these three categories (detailed in Appendix \ref{sec:im_sorr}), significantly enhancing the stability of model evaluations.

\subsection{Training Settings}
\label{sec:trainsetttt}
We only need to train the process Latent Reward model. 
In our experiments, we train such a model using Qwen2.5-7B-Instruct. To avoid disrupting the native structure of the LLM, we adopt a generative PRM training method. Specifically, we designate two special tokens to represent the \texttt{``more potential''} and \texttt{``less potential''} labels based on Latent Reward values, and then fully reuse the training method of SFT. We train the Qwen2.5-7B-Instruct model on the collected process data for 2 epochs with a learning rate of 1e-6. More training details can be found in the Appendix \ref{appendix:details}

\begin{table*}[!t]
    \centering
    \small
    \resizebox{\textwidth}{!}{
    \begin{tabular}{lccccccccc}
    \toprule
     \textbf{Models}  & \textbf{Strategy} & \textbf{Invocation Form}  & \textbf{I1 Ins.}  & \textbf{I1 Cat.} & \textbf{I1 Tool.}& \textbf{I2 Ins.} & \textbf{I2 Cat.} & \textbf{I3 Ins.} & \textbf{Avg}\\
     \midrule
     \rowcolor[gray]{0.9} 
     \multicolumn{10}{c}{\textbf{Open-Source LLMs}} \\
    \multirow{2}{*}{ToolLLaMA-v2 (7B)} 
        & CoT & \multirow{2}{*}{JSON} & 32.06 & 37.70 & 36.01 & 29.35  & 38.00 & 29.41 & 33.39 \\
        & DFSDT & & 46.56 & 55.74 & 51.27 & 49.46 & 60.50 & 55.89 & 53.24  \\ 
        \midrule
        TP-LLaMA (7B) & DFSDT & JSON & 27.00 & 48.00 & 37.00 & 35.00 & 36.00 & 35.00 & 36.00 \\
       StepTool (7B) & DFSDT & JSON & 44.27 & 47.54 & 42.80 & 42.39 & 43.00 & 44.12 & 44.02 \\
     \midrule
     \multirow{3}{*}{Qwen2.5-Instruct-7B} 
        & CoT & JSON & 56.11 & 58.20 & 51.26 & 54.89  & 61.00 & 44.12 & 54.26 \\
        & DFSDT & JSON & 61.07 & 63.11 & 62.18 & 54.89 & 66.00 & 55.88 & 60.52 \\ 
        & {\textbf{CodeTool}} & Code & \textbf{63.74} & \underline{68.44} & \textbf{65.68} & \underline{55.98} & \textbf{72.50} & 
         \underline{58.82} & \underline{64.19}  \\
    \midrule
    CodeLlama-7B & \multirow{2}{*}{\textbf{CodeTool}} & \multirow{2}{*}{Code} & 45.80 & 46.72 & 42.37 & 45.65 & 47.50 & 35.29 & 43.89  \\
    Qwen2.5-Coder-7B & & & \underline{62.59} & \textbf{74.59} & \underline{63.98} & \textbf{67.93} & \underline{70.00} & \textbf{79.41} & \textbf{69.75}  \\
     \midrule
     \rowcolor[gray]{0.9}
     \multicolumn{10}{c}{\textbf{Closed-Source LLMs}} \\
    \multirow{3}{*}{\rotatebox{0}{GPT-3.5-Turbo-0613}} & CoT & \multirow{2}{*}{JSON} & 47.71 & 45.77 & 54.66 & 38.59 & 43.00 & 44.12 & 45.64  \\
     & DFSDT  & & 59.54 & 62.29 & 66.10 & 52.72 & 62.50 & 70.58 & 62.34 \\
     \cmidrule(r){2-10}
    & \textbf{CodeTool} & Code & 59.92 & 59.84 & 53.39 & 40.21 & 53.00 & 55.88 & 53.71 \\
    \midrule
    \multirow{3}{*}{\rotatebox{0}{GPT-4-Turbo-Preview}} & CoT & \multirow{2}{*}{JSON} & 51.15 & 64.34 & 55.84 & 55.43 & 58.50 & 70.59 & 59.32 \\
     & DFSDT  & & 59.16 & 61.06 & 47.48 & 62.50 & 65.50 & 76.47 & 62.03 \\
     \cmidrule(r){2-10}
     & \textbf{CodeTool} & Code & \textbf{62.97} & \textbf{76.22} & \textbf{69.49} & \textbf{65.76} & \textbf{69.50} & \textbf{82.35} & \textbf{71.05}  \\
    \bottomrule
    \end{tabular}}
    \caption{Performance comparison in terms of SoPR between CodeTool and baselines. The best performances in open-source LLMs are highlighted in \textbf{bold}, while suboptimal ones are marked with \underline{underline}. We reproduce StepTool on ToolLLaMA-v2 using the released code. All results are assessed on the filtered StableToolBench test set using the improved SoPR evaluation prompt (refer to Section \ref{sec:baseline_metric} for details about the evaluation prompt).}
    \label{tab:main}
\end{table*}

\section{Experimental Results and Analyses}
\label{sec:main_res_ans}

\subsection{Main Results}
Table \ref{tab:main} presents the performance comparison of CodeTool with baselines on both open-source and closed-source LLMs.
Below are some key observations drawn from the results:
(\textbf{1}) For open-source LLMs, when Qwen2.5-7B-Instruct is used as the code generation model, CodeTool consistently outperforms both the CoT and DFSDT strategies.
(\textbf{2}) For open-source Coder LLMs, while CodeLlama-7B exhibits lower performance compared to ToolLLaMA when the DFSDT strategy is adopted, the more capable Qwen2.5-Coder-7B-Instruct, which possesses stronger coding and instruction-following abilities, achieves significantly higher SoPR scores on the test set, demonstrating the effectiveness of the CodeTool framework. Although Qwen2.5-7B-Instruct and Qwen2.5-Coder-7B-Instruct perform comparably across different test subsets, with each showing advantages in certain areas, Qwen2.5-Coder-7B-Instruct ultimately offers more consistent and overall better performance.
(\textbf{3}) For closed-source LLMs, both GPT-3.5-Turbo-0613 and GPT-4-Turbo-Preview demonstrate comparable performance on the SoPR metric when employing the DFSDT strategy with JSON-format-based tool invocations. Notably, GPT-4-Turbo-Preview, with its superior code generation capabilities, further elevates the SoPR metric when paired with CodeTool, enhancing the performance of Qwen2.5-Coder-7B-Instruct by 1.86\%.
(\textbf{4}) The results on both open-source and closed-source LLMs suggest that the more advanced the model's coding capabilities, the better its tool invocation performance in terms of the SoPR metric when paired with our CodeTool.

\subsection{Generalization Performance on RestBench-TMDB}

\begin{table}[!t]
    \centering
    \small
    \resizebox{1\columnwidth}{!}{
    \begin{tabular}{lcc}
    \toprule
     \multirow{2}{*}{\textbf{Methods}}  & \multicolumn{2}{c}{\textbf{RestBench-TMDB}} \\
     \cmidrule(r){2-3}
     & \textbf{Success (\%)} & \textbf{Path Rate (\%)} \\
     \midrule
     \rowcolor[gray]{0.9} 
     \multicolumn{3}{c}{\textbf{GPT-3.5-Turbo-0613}} \\
     \midrule
    ReAct \citep{react} & 61.00 & 77.13 \\
    RestGPT \citep{restgpt} & 65.00 & 77.49 \\
    CodeAct \citep{codeact} & 63.00 & 80.91 \\ 
    ToolLLaMA \citep{toolllama} & 72.00 &  78.29 \\
    ATC \citep{ACT} & 89.00 & 84.71 \\
    \midrule
    \textbf{CodeTool} &  \textbf{92.00} & \textbf{91.15} \\
    \bottomrule
    \end{tabular}}
    \caption{Performance comparison in terms of success rate and correct path rate on RestBench-TMDB.}
    \label{tab:restdb}
\end{table}

To evaluate the generalization capability of the CodeTool framework, we conduct additional experiments on another widely recognized benchmark, RestBench-TMDB \cite{restgpt}, which contains 100 tasks involving 54 tools designed for movie-related scenarios. Notably, we do not retrain the process Latent Reward model for this benchmark; instead, we utilize the model previously trained for StableToolBench directly.
We adopt the Success Rate and Correct Path Rate metrics provided by RestBench-TMDB for evaluation. Success Rate relies on human assessment to ascertain whether the model's output successfully fulfills the user query, while Correct Path Rate measures the proportion of ground truth tools in model-generated tool invocations. The results, as shown in Table \ref{tab:restdb}, indicate that CodeTool achieves the best performance, with a Success Rate of 92\% and a Pass Rate of 91.15\%, surpassing ATC by 3.37\% and 7.60\%, respectively, which involves writing complete Python code to solve user queries.

\begin{table*}[!t]
    \centering
    \small
    \resizebox{\textwidth}{!}{
    \begin{tabular}{lccccccc}
    \toprule
     \multirow{2}{*}{\textbf{Methods}}  & \multicolumn{7}{c}{\textbf{Solvable Pass Rate (SoPR, \%)}} \\
     \cmidrule(r){2-8}
     & \textbf{I1 Ins.}  & \textbf{I1 Cat.} & \textbf{I1 Tool.}& \textbf{I2 Ins.} & \textbf{I2 Cat.} & \textbf{I3 Ins.} & \textbf{Avg}\\
     \midrule
    Qwen2.5-Coder-7B + CodeTool & 62.59 & \textbf{74.59} & \textbf{63.98} & \textbf{67.93} & 70.00 & \textbf{79.41} & \textbf{69.75}  \\
    - w/o On-the-spot Reward & 61.45 & 72.95 & 61.86 & 60.32 & 68.50 & 70.83 & 65.99 \\ 
    - w/o Latent Reward & \textbf{67.94} & 68.03 & 62.71 & 62.50 & \textbf{73.00} & 58.33 & 65.41 \\
    \bottomrule
    \end{tabular}}
    \caption{Ablation study on the two types of process rewards within CodeTool from the perspective of SoPR.}
    \label{tab:ablation_SoPR}
\end{table*}

\begin{table*}[!t]
    \centering
    \small
    \resizebox{\textwidth}{!}{
    \begin{tabular}{lccccccc}
    \toprule
     \multirow{2}{*}{\textbf{Methods}}  & \multicolumn{7}{c}{\textbf{Successful Code Execution Proportion (SCEP, \%)}} \\
     \cmidrule(r){2-8}
     & \textbf{I1 Ins.}  & \textbf{I1 Cat.} & \textbf{I1 Tool.}& \textbf{I2 Ins.} & \textbf{I2 Cat.} & \textbf{I3 Ins.} & \textbf{Avg}\\
     \midrule
    Qwen2.5-Coder-7B + CodeTool & \textbf{85.33} & \textbf{84.62} & \textbf{82.47} & \textbf{87.14} & \textbf{86.61} & \textbf{95.00} & \textbf{86.86} \\
    - w/o On-the-spot Reward & 66.45 & 71.03 & 69.20 & 71.10 & 58.63 & 80.36 & 69.46 \\ 
    - w/o Latent Reward & 84.46 & 84.17 & 80.93 & 84.48 & 86.34 & 91.67 & 85.34 \\
    \bottomrule
    \end{tabular}}
    \caption{Ablation study on the two types of process rewards within CodeTool from the perspective of SCEP.}
    \label{tab:SCEP}
\end{table*}

\subsection{Analyses}

\paragraph{Ablation Study I: Impact of Two Types of Process Rewards on SoPR}
To evaluate the contribution of the two types of process rewards in CodeTool, we conduct an ablation study by constructing two variants: \textit{- w/o On-the-spot Reward}, where the On-the-spot Reward is set to 0 regardless of the success of code execution at the current step, and \textit{- w/o Latent Reward}, where the Latent Reward is set to 0, meaning that the new code generation step only aims for a direction of correct code execution. If multiple directions for correct execution exist at the current step, one is randomly selected.
As shown in Table \ref{tab:ablation_SoPR}, the removal of On-the-spot Reward and Latent Reward results in a decrease in the average SoPR to 65.99 and 65.41, respectively. These findings suggest that both types of process rewards positively impact programmatic tool invocation within CodeTool.
However, removing Latent Reward leads to an increase in SoPR on \textit{I1\_Ins} and \textit{I2\_Cat}. 
The reason may be that the queries in \textit{I1\_Ins} and \textit{I2\_Cat} are relatively simple compared to other subsets. Consequently, the guidance provided by the On-the-spot Reward is sufficient to address these queries, while the Latent Reward from the PRM may introduce additional biases, resulting in weaker performance.
Nevertheless, the average SoPR for the \textit{- w/o Latent Reward} setting is 65.41, which is much lower than the 69.75 achieved when Latent Reward is included. Therefore, it can be concluded that Latent Reward retains significant importance within CodeTool.

\paragraph{Ablation Study II: Impact of Two Types of Process Rewards on Successful Code Execution Proportion}
To further investigate the role of the process reward system within the CodeTool framework, we conduct another ablation study to evaluate the impact of the two types of process rewards on the Successful Code Execution Proportion (\textbf{SCEP}). This metric assesses the overall correctness of stepwise code execution and is defined as follows:
\begin{equation}
SCEP = \frac{\sum_{i=1}^{6} \sum_{j=1}^{P_i} \sum_{k=1}^{S_{i, j}} C_{i, j,k}}{\sum_{i=1}^{6} \sum_{j=1}^{P_i} S_{i, j}},
\end{equation}
where \( P_i \) represents the number of queries in subset \( i \) (for \( i = 1, 2, \dots, 6 \)), \( S_{i, j} \) denotes the number of coding steps required to solve the query \( j \) in subset \( i \), and \( C_{i, j,k} \) is an indicator of whether the coding step \( k \) in addressing query \( j \) (from subset \( i \)) is correctly executed, with \( C_{i, j,k} = 1 \) if the step is correctly executed and \( C_{i, j,k} = 0 \) otherwise.

As shown in Table \ref{tab:SCEP}, removing On-the-spot Reward significantly lowers the average SCEP to 69.46\%, indicating its crucial role in guiding the CodeTool framework to generate accurate code. This is expected, as the On-the-spot Reward ensures that the LLM's code generation is directed towards correct execution. Additionally, the removal of the Latent Reward also results in a decrease in SCEP compared to when it is retained. Although the On-the-spot Reward provides immediate guidance, the Latent Reward contributes to long-term accuracy in code generation, highlighting its importance in the overall process.

\begin{figure}[!t]
    \centering
    \includegraphics[width=\linewidth]{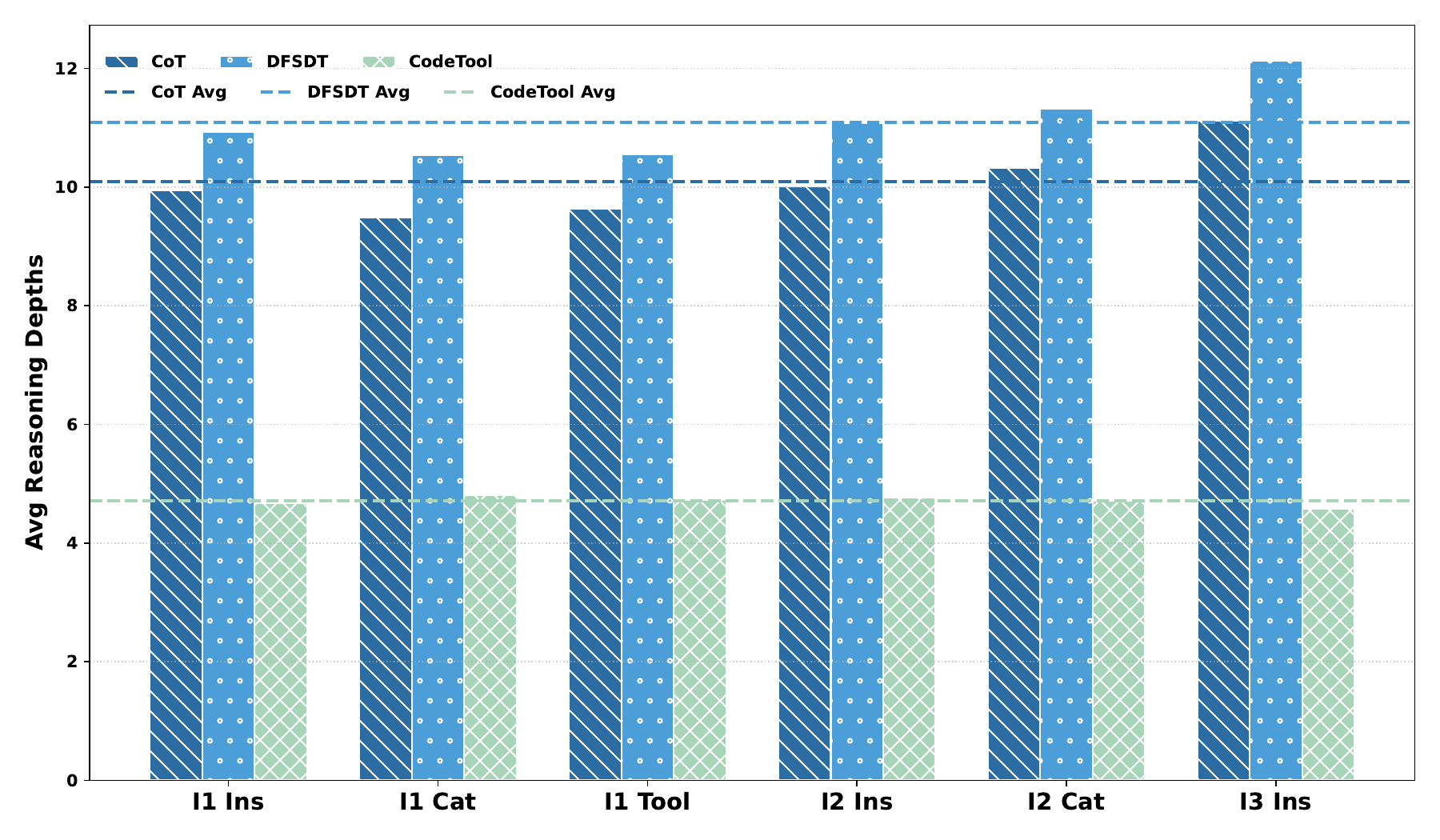}
    \caption{Average reasoning depths of three different strategies across six test subsets in StableToolBench.}
    \label{fig:depth}
\end{figure}

\begin{figure}[!t]
    \centering
    \includegraphics[width=\linewidth]{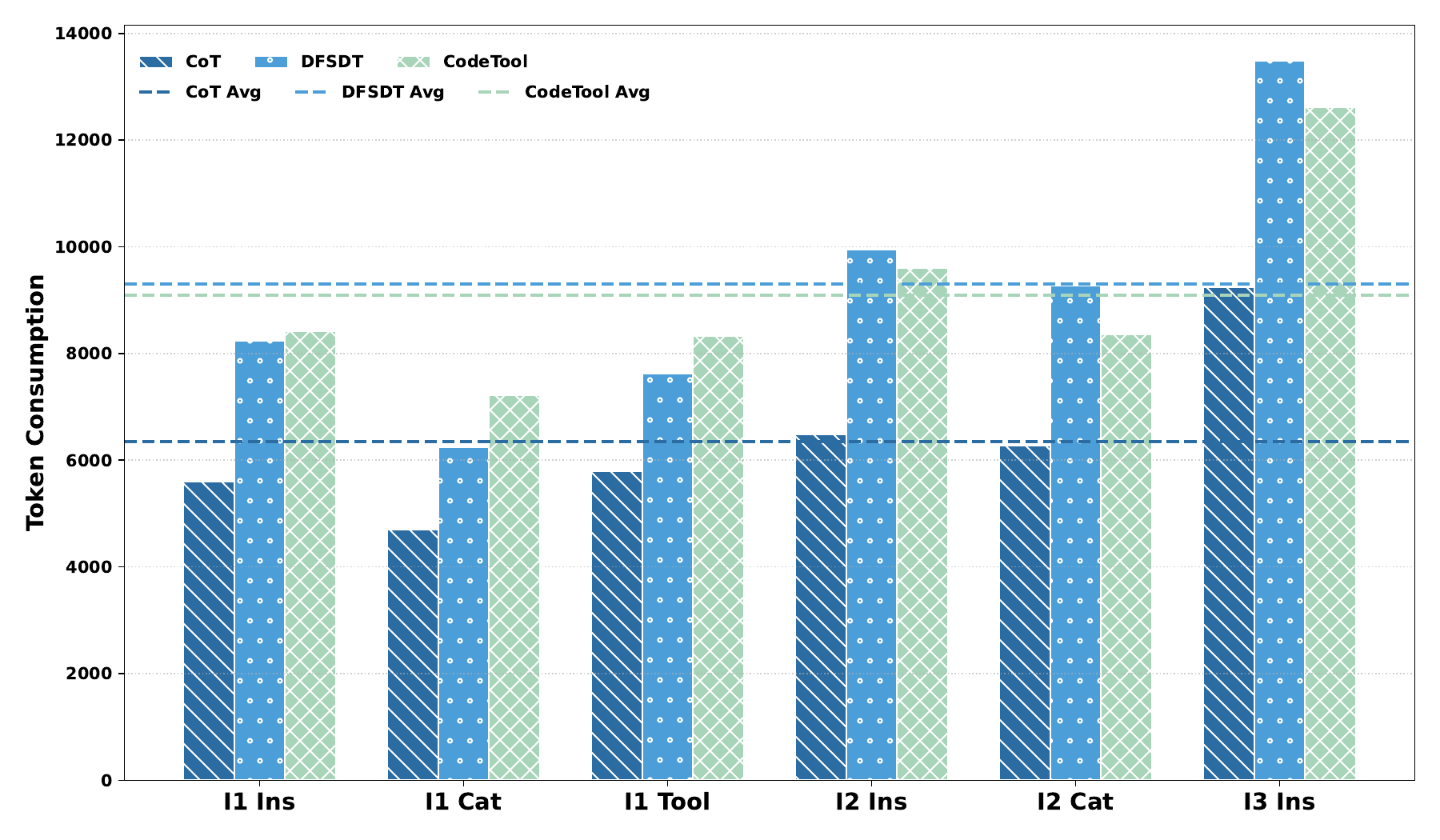}
    \caption{Token consumption of three different strategies across six test subsets in StableToolBench.}
    \label{fig:token_cosume}
\end{figure}

\paragraph{Efficiency Comparison with JSON-based Tool Invocation}
To further investigate the differences in reasoning efficiency between code-based and JSON-based methods, particularly in terms of reasoning depth and token consumption, we conduct additional comparative experiments on I2-Category. We use Qwen2.5-7B-Instruct as the code generation model. As shown in Figure \ref{fig:depth}, in terms of average reasoning depth, the CodeTool framework, compared with JSON-based tool invocation methods that employ CoT and DFSDT strategies, enables both tool invocation and response acquisition within a single interaction step, thereby effectively reducing the number of interactions by nearly half. 
As shown in Figure \ref{fig:token_cosume}, the CodeTool framework maintains a comparable level of token consumption to the DFSDT strategy, while outperforming both DFSDT and CoT in terms of SoPR, further underscoring the effectiveness of our approach.

\begin{figure}[!t]
    \centering
    \includegraphics[width=\linewidth]{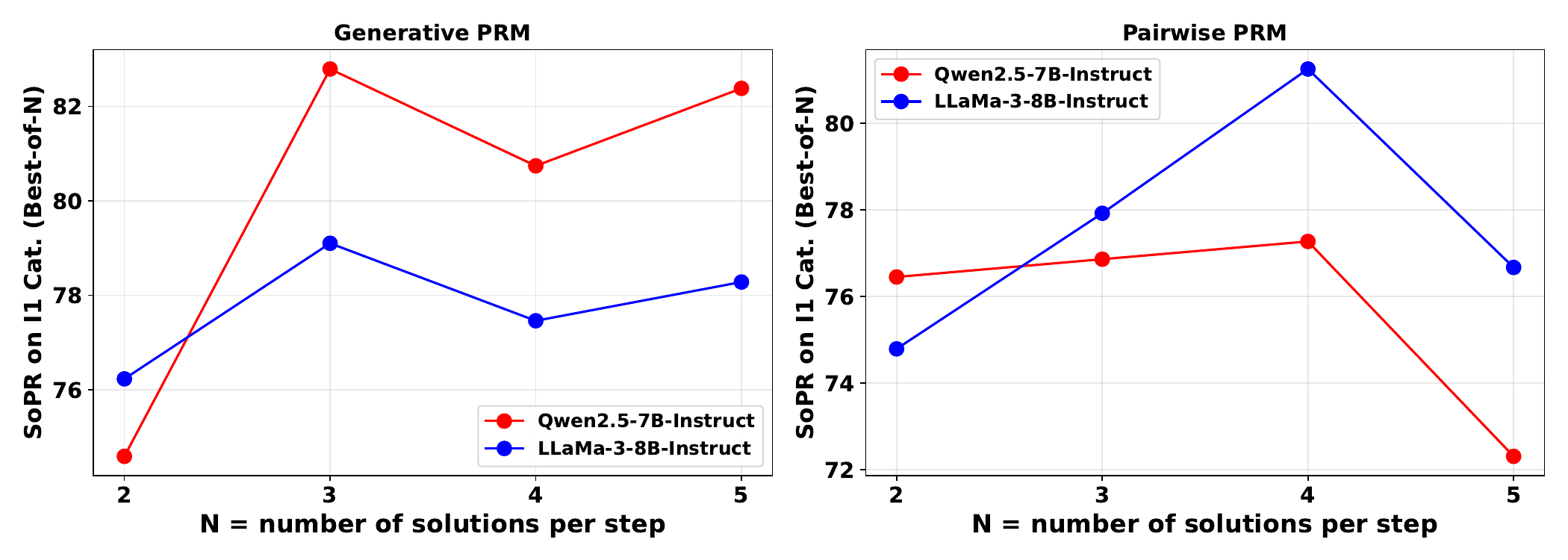}
    \caption{Performance with varying PRM training methods and number of candidate actions on I1 Category.}
    \label{fig:line_chart}
\end{figure}

\paragraph{Performance with Varying PRM Training Methods and Number of Candidate Actions}
As outlined in Section \ref{sec:trainsetttt}, we train the Generative PRM by designating two special tokens to represent the \texttt{``more potential''} or \texttt{``less potential''} labels based on Latent Reward values through a generative approach. For comparison, we also train a Pairwise PRM by augmenting the LLM backbone with an additional linear head, enabling it to classify the intermediate process data as either \texttt{``more potential''} or \texttt{``less potential''}.
Apart from Qwen2.5-7B-Instruct, we also train LLaMA-3-8B-Instruct \cite{llama3modelcard} as the PRM. As shown in Figure \ref{fig:line_chart}, increasing the number of candidate actions from 2 to 3 consistently improves SoPR across both PRM training methods and models. However, further increasing $N$ beyond 3 tends to degrade performance, indicating a trade-off between diversity and quality in candidate selection. Under the Generative PRM setting, Qwen2.5-7B-Instruct consistently outperforms LLaMA-3-8B-Instruct, especially at \( N = 3\) and \( N = 5\). In contrast, under the Pairwise PRM setting, LLaMA-3-8B-Instruct achieves the highest SoPR at \( N = 4\), surpassing Qwen2.5-7B-Instruct. These observations highlight the importance of both the model architecture and the training paradigm in determining the effectiveness of PRM, and suggest that model-specific tuning of $N$ and PRM strategy may be necessary for optimal performance.


\section{Conclusion}
In this paper, we introduce CodeTool, a stepwise code generation framework based on process supervision. By leveraging code as a naturally verifiable format, we obtain an On-the-spot Reward to reflect step correctness and train a PRM on fully automated process data to assign a Latent Reward, which measures the potential of each step toward overall task completion. At each inference step, LLMs follow the reasoning path with maximal cumulative reward of On-the-spot Reward and Latent Reward. Extensive experiments conducted on StableToolBench and RestBench-TMDB validate the effectiveness of the CodeTool framework.  


\section*{Limitations}

Despite the superiority of the proposed CodeTool framework, its performance is, to some extent, influenced by the code generation capabilities of the underlying LLM. A model with high proficiency in generating accurate and efficient code will naturally enhance the performance of CodeTool. Conversely, models with less advanced coding abilities may not fully exploit the potential of this framework. In addition, previous work shows that the performance of the PRM for Latent Reward is closely tied to the quality of the collected process data, particularly the accuracy of the Latent Reward values \cite{qwen2math-prm}. Given that we rely on sampling methods to estimate these values and subsequently use them to train the PRM, there is a potential for suboptimal performance if the estimated values are not sufficiently accurate.





\section*{Ethics Statement}
The research conducted in this paper aims at enhancing programmatic tool invocation of LLMs
via process supervision. Throughout the course of this research, we have rigorously adhered to ethical standards to uphold the integrity and validity of our work. All tools (APIs) utilized in this study are sourced from publicly available platforms, ensuring full transparency and reproducibility in our experimental procedures. Moreover, we have taken great care to ensure that our research does not cause harm to individuals or groups, and we have committed to avoiding any forms of deception or misuse of information in the course of our study.

\bibliography{references}
\bibliographystyle{acl_natbib}

\appendix
\clearpage 
\section{Instruction for Stepwise Code Generation}
\label{appendix:instruction}
\begin{figure*}[h]
    \centering
    \includegraphics[width=1\linewidth]{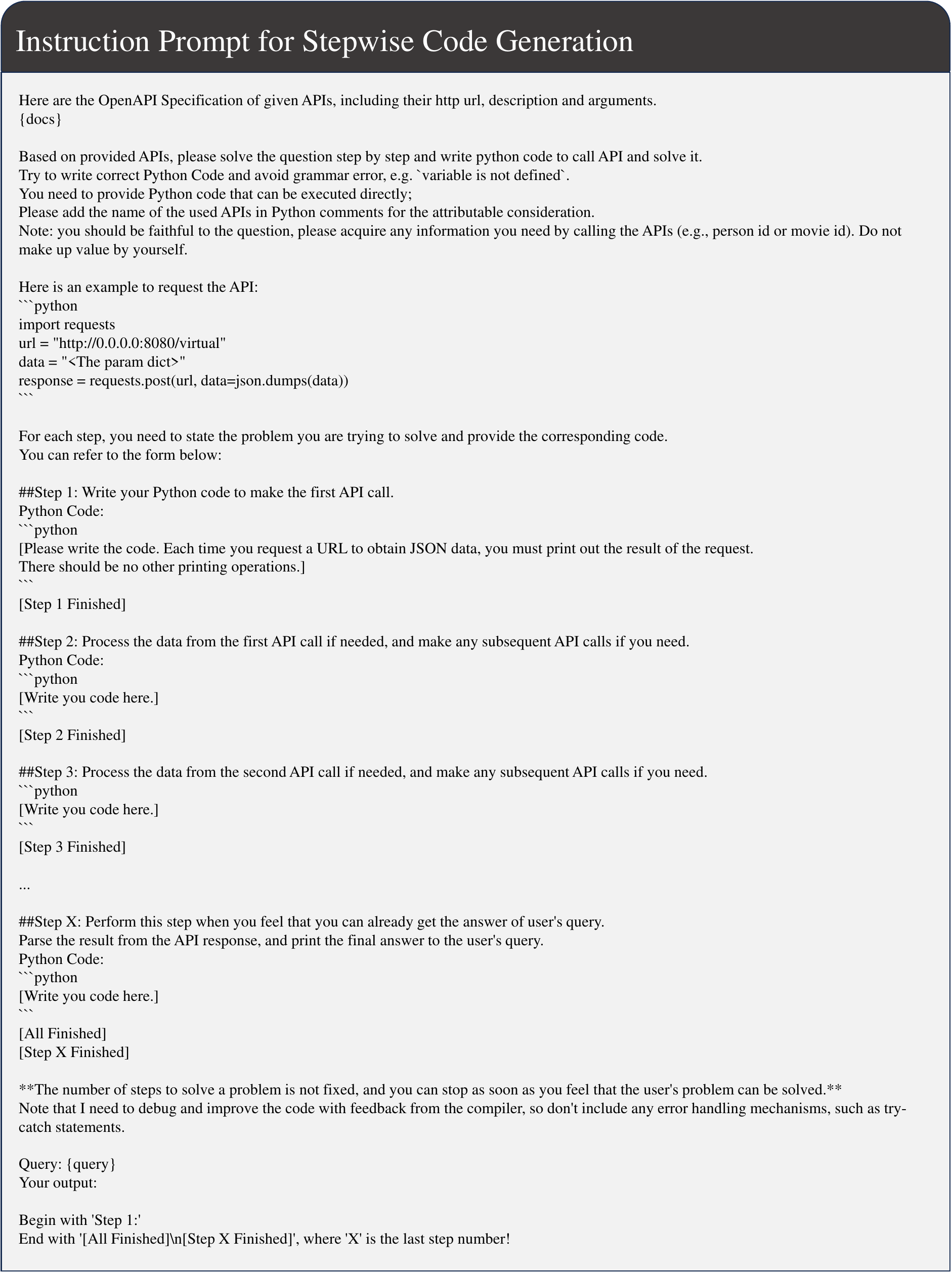}
    \caption{A reference prompt for stepwise code generation.}
    \label{fig:prompt}
\end{figure*}

A reference prompt designed to guide LLMs in performing code generation step-by-step is illustrated in Figure \ref{fig:prompt}. This prompt is structured to facilitate incremental reasoning by the model, providing clear instructions.

\section{Improved SoPR Evaluation Prompt}
\label{sec:im_sorr}
To enhance the stability of model evaluations, we introduce clearer criteria for assessing \texttt{``Solved''}, \texttt{``Unsolved''}, or \texttt{``Unsure''} cases for the SoPR metric. The improved evaluation prompt is shown in Table \ref{tab:prompt}.

\begin{table*}[!ht]
    \centering
    \small
    \begin{tabular}{l}  
    \toprule
     \rowcolor[gray]{0.9} 
     \multicolumn{1}{c}{\textbf{SoPR Evaluation Prompt in StableToolBench}} \\
     \midrule
     Giving the query and answer, you need to give \texttt{answer\textunderscore status} of the answer by following rules: \\
     1. If the answer is a sorry message or not a positive/straight response for the given query, return \texttt{"Unsolved"}. \\
     2. If the answer is a positive/straight response for the given query, you have to further check. \\
     2.1 If the answer is not sufficient to determine whether it solves the query or not, return \texttt{"Unsure"}. \\
     2.2 If you are confident that the answer is sufficient to determine whether it solves the query or not, return \texttt{"Solved"} or \texttt{"Unsolved"}. \\
     \\
     Query: \{query\} \\
     \\
     Answer: \{answer\} \\
     \\
     Now give your reason in \texttt{"content"} and \texttt{"answer\textunderscore status"} of JSON to \texttt{"check\textunderscore answer\textunderscore status"}. \\
     \midrule
     \rowcolor[gray]{0.9} 
     \multicolumn{1}{c}{\textbf{Improved SoPR Evaluation Prompt in CodeTool}} \\
     \midrule
     Giving the query and answer, you need to give \texttt{answer\textunderscore status} of the answer by following rules: \\
     1. If the answer doesn't contain any information that is helpful for answering the user's query, return \texttt{"Unsolved"}. \\
     2. If the answer is a positive/straight response for the given query, you have to further check. \\
     2.1 If the answer is not sufficient to determine whether it solves the query or not, return \texttt{"Unsure"}. \\
     2.2 If the answer solves part of the query or does not fully answer the query, return \texttt{"Unsure"}. \\
     2.3 If the answer is sufficient to solve the query, return \texttt{"Solved"}. \\
     \\
     Query: \{query\} \\
     \\
     Answer: \{answer\} \\
     \\
     Now give your reason in \texttt{"content"} and \texttt{"answer\textunderscore status"} of JSON to \texttt{"check\textunderscore answer\textunderscore status"}. \\
    \bottomrule
    \end{tabular}
    \caption{We have refined the criteria for each category in the SoPR assessment prompt, making the SoPR assessment more stable.}
    \label{tab:prompt}
\end{table*}

\section{StableToolBench Test Set Filtering Rules}
Rules for filtering the test set of StableToolBench are shown in Table \ref{table:rules}.
\label{appendix:rules}

\begin{table*}[!h]
    \centering
    \small
    \begin{tabular}{p{16cm}}  
    \toprule
     \rowcolor[gray]{0.9} 
     \multicolumn{1}{c}{\textbf{Rules for Filtering the Test Set of StableToolBench.}} \\
     \midrule
        1. The parameters for requesting the API in StableToolBench are inconsistent with those in the given API documentation, resulting in the inability to request the API. \\[0.5ex]
        \\
        2. When the LLM provides the same API and request parameters as those in the StableToolBench experiments, the response in StableToolBench can solve the problem. However, the content in the cache is just a piece of text and cannot solve the problem. \\[0.5ex]
        \\
        3. Even if the LLM provides the same API and request parameters as those in the StableToolBench experiments, the request fails to be fulfilled due to the lack of cache. Moreover, the real ToolBench API either does not exist, or requires a subscription, or there is no access permission or returns a piece of text that cannot solve the query. \\
    \bottomrule
    \end{tabular}
    \caption{Rules for Filtering the Test Set of StableToolBench.}
    \label{table:rules}
\end{table*}

\section{Total Number of APIs Involved in the Filtered Test Set}
The total number of APIs involved in the filtered test set is shown in Table~\ref{table:apis}.
\begin{center}
\begin{minipage}{\linewidth}
    \centering
    \small
    \resizebox{\linewidth}{!}{
    \begin{tabular}{lcccccc}
    \toprule
    \textbf{} & \textbf{I1-I} & \textbf{I1-C} & \textbf{I1-T} & \textbf{I2-I} & \textbf{I2-C} & \textbf{I3-I} \\
    \midrule
    \textbf{Task} & 131 & 122 & 119 & 92 & 100 & 17 \\
    \textbf{Candidate API} & 631 & 271 & 371 & 536 & 341 & 31 \\
    \textbf{Relevant API} & 288 & 171 & 199 & 197 & 172 & 19 \\
    \bottomrule
    \end{tabular}
    }
    \captionof{table}{The total number of APIs involved in the filtered test set. C, I, T stand for the Category, Instruction, and Tool subgroup of the test set.}
    \label{table:apis}
\end{minipage}
\end{center}

\section{Details of PRM Training}
\label{appendix:details}
In the process data collection phase, we draw inspiration from the sampling strategy of MCTS. However, considering the high time and computational cost associated with performing multiple rollouts in MCTS, we adopt a simplified implementation. As described in Section \ref{sec:3.4} of the paper, to balance computational resources and inference efficiency, we employ a Depth-First Search (DFS) algorithm. Specifically, during the code generation phase at each step of tool invocation, we perform two samples for each node's next action, thereby expanding a binary tree-like search space for code generation paths. For each node in the search tree, we first compute its raw Latent Reward using Equation \ref{equa:raw_latent_reward}, where a key challenge lies in determining whether the current reasoning path successfully solves the user's query. In the scenario of code-based tool invocation, we adopt the practice of \citet{ACT} by explicitly instructing the model in the prompt to output a complete \textit{Final Answer} via a print statement at the final step of reasoning. In relatively straightforward tool-invocation scenarios (e.g., numerical queries), the model typically follows this instruction well, accurately providing a complete \textit{Final Answer}. However, in more complex tasks, it is difficult for the model to directly generate a complete answer in the final step. Thus, we implement a more flexible strategy: after each tool invocation, the model immediately prints out the result returned from that request, as specified explicitly in Figure \ref{fig:prompt}. Subsequently, we pair the model's thought at each reasoning step with the corresponding tool response, and combine these pairs with the original user query to prompt the LLM to reorganize them into a coherent natural-language text as the \textit{Final Answer}. After obtaining the \textit{Final Answer}, we apply the same evaluation prompt used in the SoPR experiment (detailed in Table \ref{tab:prompt}). Based on how well the model-generated \textit{Final Answer} addresses the user's original query, LLM classifies each answer into one of three categories: \texttt{``Solved''}, \texttt{``Unsure''}, or \texttt{``Unsolved''}. Accordingly, we automatically assign its final Latent Reward using Equation \ref{equa:latent_reward}. Leveraging these reward signals, we further collect process data with varying Latent Reward values from every level of the search tree, thereby constructing training samples for the PRM that capture the differences in potential among various paths.

In the training phase of the PRM, to avoid interfering with the structure of the LLM, we follow a generative training strategy similar to \citet{mathshepherd}. Specifically, for two child nodes sampled from the same parent node, we assign labels \texttt{``more potential''} and \texttt{``less potential''} based on their Latent Reward values (e.g., using two special tokens \texttt{``yes''} and \texttt{``no''} as supervision signals) for supervised fine-tuning.

During inference, to estimate the Latent Reward of the current step, we compute the logits of the \texttt{``yes''} and \texttt{``no''} tokens output by the model and normalize them using the following formula:
\begin{equation}
    \text{Latent Reward} = \frac{\text{logit}_{\text{yes}}}{\text{logit}_{\text{yes}} + \text{logit}_{\text{no}}}.
\end{equation}

\section{Analysis of Rewards Conflict}
To further investigate the Reward Conflict phenomenon during inference, we conduct inference using Qwen2.5-7B-Instruct as the code generation model on the I2-Category subset, with the number of candidate actions set to 2. In total, we collect 497 pairs of intermediate reasoning data. The detailed reward distribution is shown in Table~\ref{tab:reward_conflict_cases}. 

Case 1 refers to \textbf{Both candidates have an On-the-spot Reward of 1 (i.e., both are executable);}

Case 2 refers to \textbf{Both candidates have an On-the-spot Reward of 0 (i.e., neither candidate is executable);}

Case 3 refers to \textbf{One candidate node is executable while the other is not, and the executable node has a higher Latent Reward;}

Case 4 refers to \textbf{One candidate is executable, but has a lower Latent Reward than the non-executable one (i.e., the reward conflict scenario.)}

\begin{center}
    \begin{minipage}{\linewidth}
    \centering
    \begin{tabular}{lcc}
    \toprule
    \textbf{Case Type} & \textbf{Number} & \textbf{Ratio} \\
    \midrule
    \textbf{Total Samples} & 497 & --- \\
    Case 1 & 363 & 73.04\% \\
    Case 2 & 59 & 11.87\% \\
    Case 3 & 41 & 8.25\% \\
    Case 4 (conflict) & 34 & 6.84\% \\
    \bottomrule
    \end{tabular}
    \captionof{table}{Distribution of case types including reward conflict cases.}
    \label{tab:reward_conflict_cases}
    \end{minipage}
\end{center}

We observe that reward conflict (Case 4) is indeed rare, accounting for only 6.84 \% of all sampled cases. This low occurrence suggests that, in most situations, the On-the-spot Reward and Latent Reward are aligned.

\section{Case Study}
Compared to methods that rely on JSON or Text format for tool invocation, CodeTool with process supervision, offers multiple advantages beyond its superiority in handling request-intensive instructions, as illustrated in Figure \ref{fig:overview}. The following are some specific cases:

\subsection{Case 1}
\begin{figure*}[t]
    \centering
    \includegraphics[width=1\linewidth]{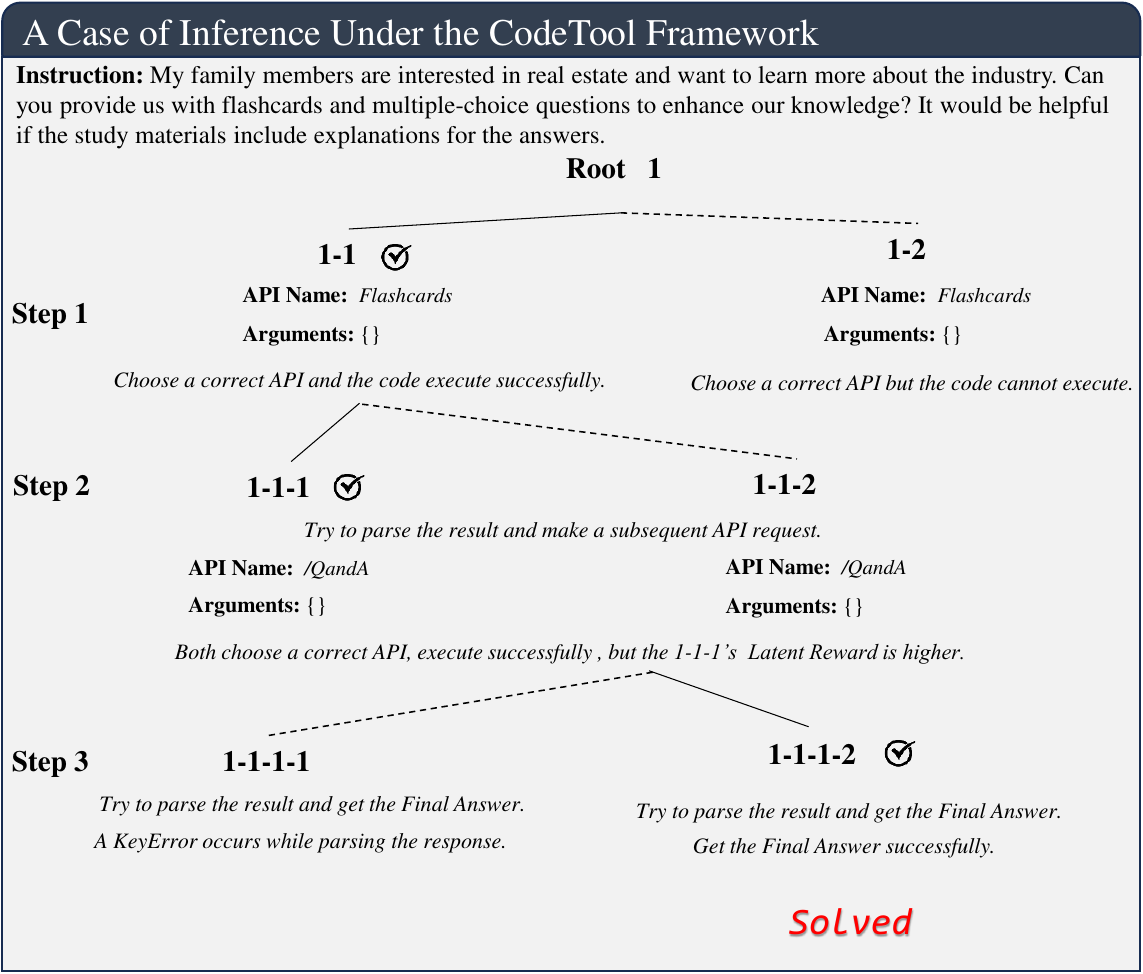}
    \caption{A case of inference under our CodeTool framework.}
    \label{fig:case1}
\end{figure*}

We use Figure \ref{fig:case1} as an example to demonstrate the reasoning process of CodeTool when sampling two candidate actions at each step, highlighting common scenarios encountered during the evaluation of each intermediate step. 

\subsection{Case 2}
\begin{figure*}[t]
    \centering
    \includegraphics[width=1\linewidth]{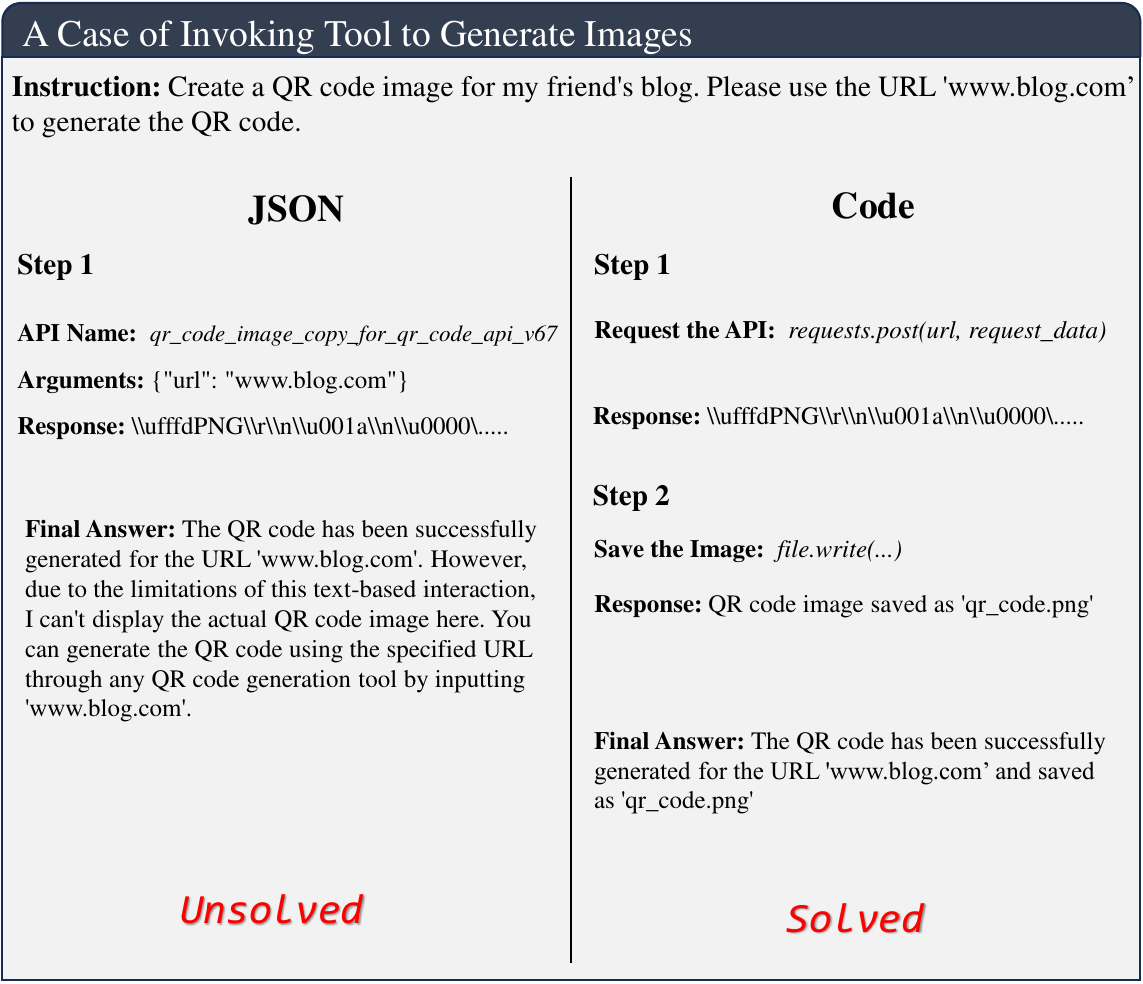}
    \caption{A case of invoking tools to generate images.}
    \label{fig:case2}
\end{figure*}

Figure \ref{fig:case2} demonstrates that when the user expects a specific tool to generate an image, although the tool invocation request based on the JSON format retrieves the image information, it fails to save the image. In contrast, the code not only successfully requests the tool but also saves the image locally, completely resolving the user's query.

\subsection{Case 3}
\begin{figure*}[t]
    \centering
    \includegraphics[width=1\linewidth]{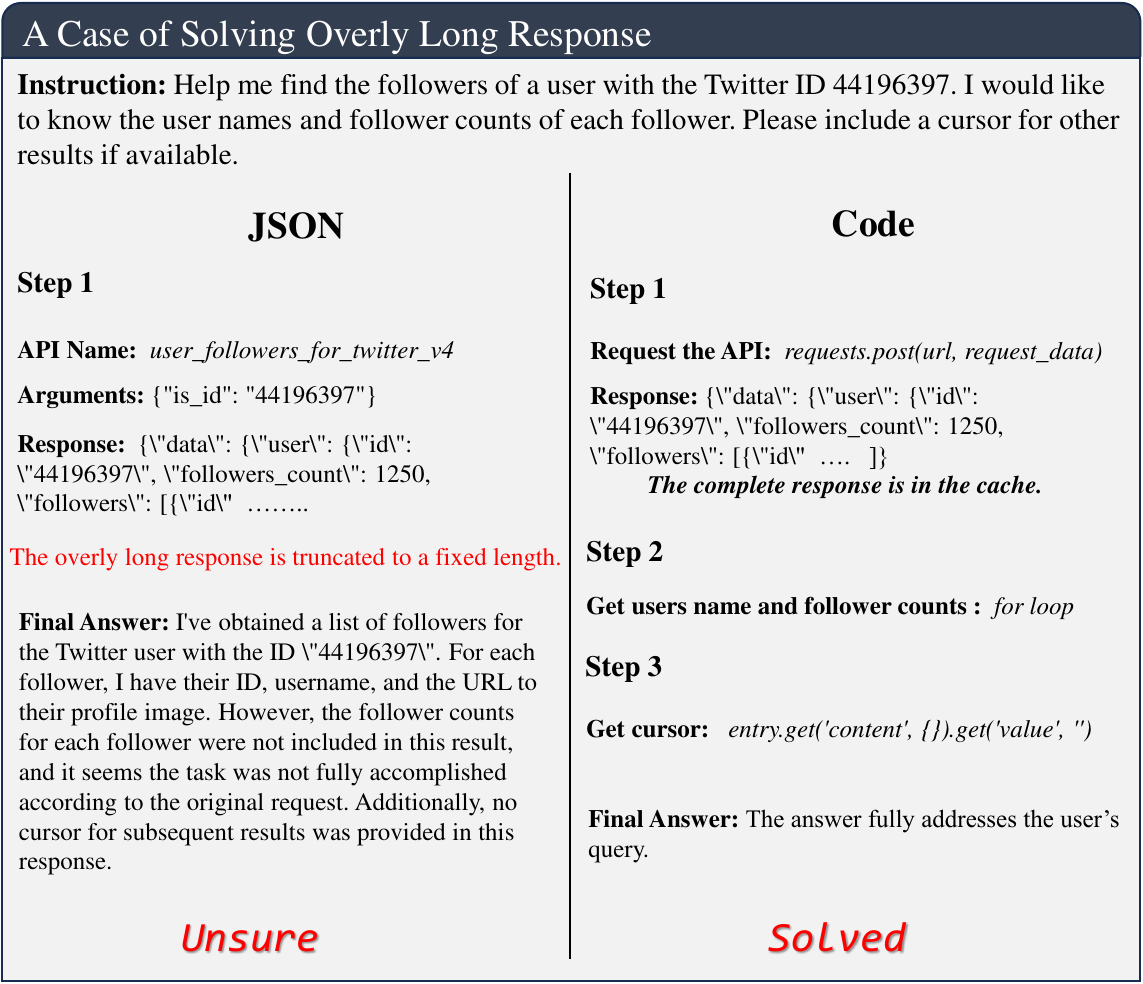}
    \caption{A case of solving overly long responses.}
    \label{fig:case3}
\end{figure*}

Figure \ref{fig:case3} demonstrates that when the tool's response is overly long and the key information is truncated, the user's query may not be resolved. However, the code stores the complete tool response in the cache, ensuring that critical information is not lost, thus better addressing the user's query.

\end{document}